\PassOptionsToPackage{table}{xcolor}
\pdftrailerid{}
\documentclass[letterpaper]{article}
\usepackage[preprint]{aaai2027}
\usepackage[hyphens]{url}
\usepackage{graphicx}
\usepackage{xcolor}
\usepackage{natbib}
\usepackage{caption}
\usepackage{booktabs}
\usepackage{tabularx}
\usepackage{amsmath}
\usepackage{amsfonts}
\usepackage{algorithm}
\usepackage{algorithmic}
\usepackage{listings}
\ifdefined\appendixonly
  \usepackage{xr}
\fi
\definecolor{skillttarow}{gray}{0.94}
\lstdefinestyle{yamlconfig}{
    basicstyle=\ttfamily\footnotesize,
    breaklines=true,
    breakatwhitespace=false,
    columns=fullflexible,
    keepspaces=true,
    showstringspaces=false,
    captionpos=t,
    frame=single,
    framesep=3pt,
    framerule=0.4pt,
    xleftmargin=2pt,
    xrightmargin=2pt
}
\lstdefinestyle{promptbox}{
    style=yamlconfig
}

\newif\ifincludeappendix
\ifdefined\mainonly
  \includeappendixfalse
\else
  \includeappendixtrue
\fi

\title{Skills on the Fly: Test-Time Adaptive Skill Synthesis for LLM Agents}
\author{
    Jingxing Wang\textsuperscript{\rm 1},
    Chenyu Zhou\textsuperscript{\rm 1},
    Zhihui Fu\textsuperscript{\rm 2},
    Jun Wang\textsuperscript{\rm 2}\corresponding,\\
    Weiwen Liu\textsuperscript{\rm 1}\corresponding,
    Weinan Zhang\textsuperscript{\rm 1}\corresponding,
    Jianghao Lin\textsuperscript{\rm 1}\corresponding
}
\affiliations{
    \textsuperscript{\rm 1}Shanghai Jiao Tong University\\
    \textsuperscript{\rm 2}OPPO Research Institute\\
    01lightyear@sjtu.edu.cn, linjianghao@sjtu.edu.cn
}

\begin{document}
\maketitle

\ifdefined\appendixonly\else

\begin{abstract}
Additional test-time compute can give LLM agents access to more past experience, yet expanding the context or adding rollouts does not necessarily yield greater agent capability. We call this challenge test-time compute-to-capability conversion and propose \emph{SkillTTA}, which retrieves task-relevant training trajectories and synthesizes a temporary skill conditioned on the visible target context for a solver with fixed parameters. To pursue a higher performance ceiling, SkillTTA further uses meta prompt optimization (MPO) to adapt the policy that writes these skills. MPO evaluates candidate prompts on paired tasks and emphasizes informative transitions. It also confines updates to benchmark-specific atomic slots, reducing the variance caused by observing each edit only indirectly through skill synthesis and solver rollout. Across ALFWorld, SpreadsheetBench, BigCodeBench, and WebShop, SkillTTA outperforms state-of-the-art reuse and optimization baselines. It attains a higher performance ceiling at lower compute cost than baseline reuse and sampling strategies.
\end{abstract}

\begin{links}
    \link{Code}{https://github.com/Hydr0pon1c/Skills-on-the-fly}
\end{links}

\section{Introduction}

Past experience can improve LLM agents at test time without changing their model weights, while agents themselves leave trajectories that record their decisions and environment feedback through rollouts, serving as a naive form of experience \citep{react2023,externalization2026,agentprotocols2025}. Compared with these verbose records, a skill is a more compressed and explicitly formatted representation of experience: it distills relevant evidence into instructions that a solver can reuse. This abstraction makes past experience easier to apply, but giving an agent access to more records does not by itself yield greater capability.

The central goal is therefore to convert additional test-time compute into agent capability, but Figure~\ref{fig:compute-conversion-challenges} shows two limitations. \textbf{First, existing reuse mechanisms provide poor compute-to-capability conversion.} Both raw retrieval and static global skills fail to realize the full value of the compute invested in constructing a trajectory pool. Raw retrieval passes irrelevant detours from the pool to the solver, whereas a global skill compresses the pool before the target task is observed and can therefore miss its contract \citep{trace2skill2026,awm2024}. Therefore, most compute is spent off-policy, with little reserved for task-conditioned adaptation at test time to handle distribution shift. \textbf{Second, optimizing compute-to-capability conversion has high variance.} SkillOpt optimizes one global skill \citep{skillopt2026}, while runtime memory-learning systems update learned utilities \citep{memrl2026}. Their changes are evaluated only after a solver rollout, often with synthesis or memory retrieval as an intermediate step. This indirect feedback makes credit assignment noisy.

\begin{figure}[!t]
    \centering
    \includegraphics[width=0.98\columnwidth]{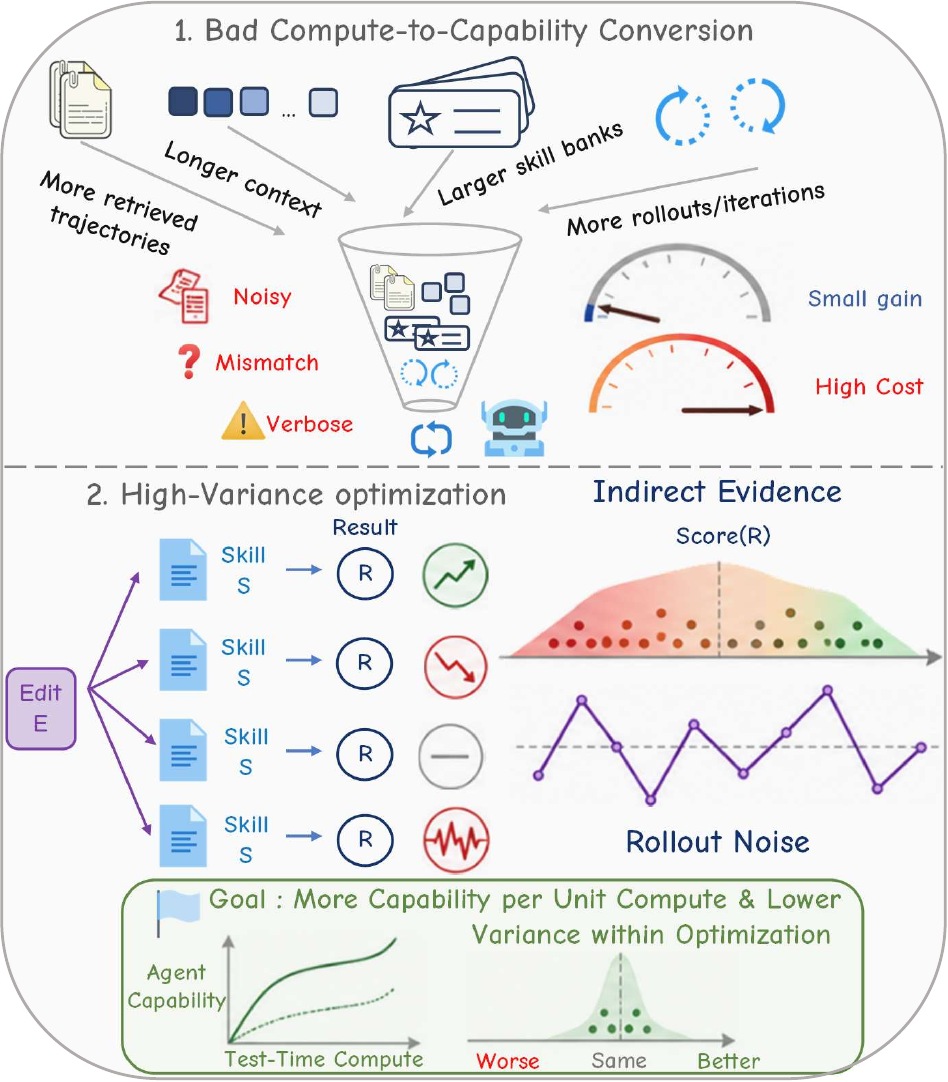}
    \caption{Two challenges in converting test-time compute into agent capability.}
    \label{fig:compute-conversion-challenges}
\end{figure}

To this end, we propose a test-time adaptive skill synthesis framework for LLM agents (e.g., \emph{SkillTTA}). First, after observing the target task, it retrieves related trajectories and turns their relevant evidence into a task-specific \texttt{SKILL.md}. A fixed solver receives this temporary skill together with the target-visible context, which excludes answers and evaluator-only information. The resulting adapter is more compact than raw trajectories and more targeted than a global skill. Second, \emph{meta prompt optimization} (MPO) adapts the synthesis policy itself and scores prompt updates on paired tasks. It reduces the remaining optimization variance through three measures: per-dataset optimization, transition-enriched sampling, and bounded atomic slot operations.

Our contributions are:
\begin{itemize}
    \item We formulate test-time compute conversion for LLM agents and study how task-conditioned synthesis turns retrieved experience into agent capability.
    \item We introduce benchmark-specific MPO for the skill-synthesis prompt. Our analysis explains how paired feedback and constrained editing control its optimization variance.
    \item Across ALFWorld, SpreadsheetBench, BigCodeBench, and WebShop, SkillTTA outperforms reuse and skill-optimization baselines. Cost and ablation analyses attribute the gains to task-conditioned skill synthesis rather than context expansion.
\end{itemize}

\section{Related Work}

\paragraph{Trajectory-to-skill synthesis.}
Trajectory-to-skill methods compress execution traces into reusable natural-language policies. Voyager and ExpeL extract skills or insights from agent experience \citep{voyager2023,expel2023}, while Trace2Skill and Agent Workflow Memory consolidate experience into static skills or workflows \citep{trace2skill2026,awm2024}. SkillTTA differs in when synthesis occurs: it retrieves evidence after observing the target task and synthesizes a temporary skill conditioned on that task's visible requirements.

\paragraph{Agent memory and retrieval.}
Memory-augmented agents retrieve experience records, reflections, or learned utilities to guide future decisions \citep{generativeagents2023,reflexion2023,memrl2026}. SkillTTA also starts with retrieval, but does not pass the retrieved records directly to the solver. Instead, it converts them into a temporary textual skill that filters irrelevant details and emphasizes constraints enforced by the evaluator.

\paragraph{Inference scaling and long-context use.}
Inference scaling can improve reasoning, although the effective allocation of compute depends on the problem and model \citep{snell2025scaling,wu2025inference}. This direction has also been extended to long-context RAG \citep{yue2025inference}, where models may use retrieved evidence unevenly across context positions \citep{liu2024lost}. SkillTTA instead uses test-time compute to turn retrieved agent trajectories into a task-conditioned skill before solver execution.

\paragraph{Test-time adaptation and prompt optimization.}
Test-time adaptation changes model behavior during evaluation through retrieved examples, retrieved skills, or parameter updates. TARSE retrieves skills and experiences for reasoning agents \citep{tarse2026}. SkillX constructs reusable skill knowledge bases, while SkillOpt treats a natural-language skill document as an optimizable external state for a frozen agent \citep{skillx2026,skillopt2026}. ProTeGi optimizes prompts with natural-language gradients and beam search \citep{pryzant2023automatic}. TextGrad treats LLM feedback as textual gradients for optimizing variables in compound AI systems, whereas MIPRO optimizes instructions and demonstrations for multi-stage LM programs without module-level labels or gradients \citep{textgrad2024,mipro2024}. SkillTTA also adapts the textual skill supplied to a fixed solver, but restricts this optimization to auditable skill-synthesis slots. Its MPO module updates the policy for writing temporary skills instead of relying on a universal template or optimizing the final solver prompt.

\section{SkillTTA}
\label{sec:method}

SkillTTA proceeds in three steps. We first introduce base test-time synthesis, then add benchmark-level MPO, and finally use variance analysis to motivate three controls for reliable optimization. Figure~\ref{fig:skilltta-overview} shows the framework.

\begin{figure*}[t]
    \centering
    \includegraphics[width=0.94\textwidth]{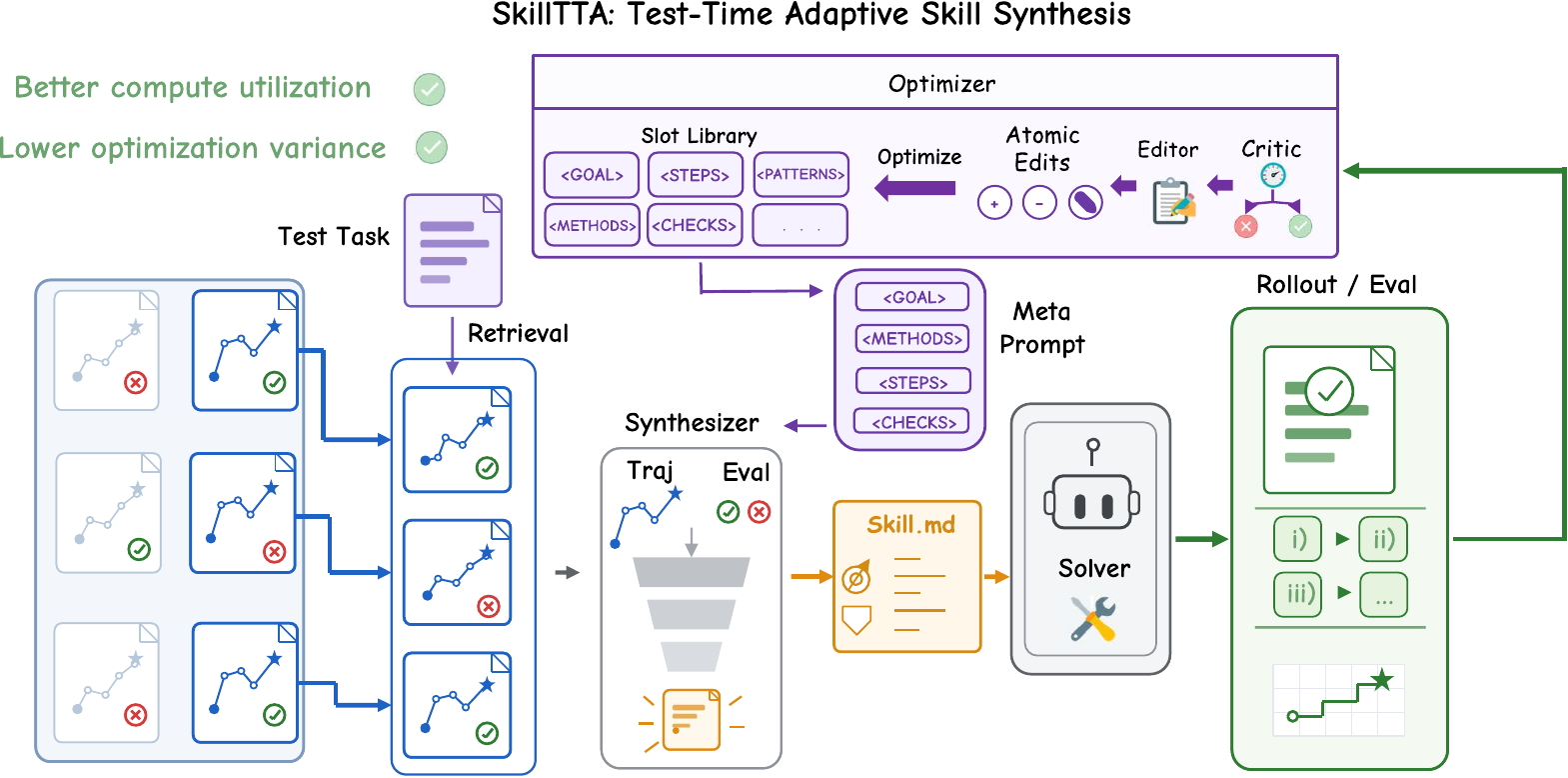}
    \caption{Overview of SkillTTA. For each test task, SkillTTA retrieves related training trajectories, synthesizes a temporary \texttt{SKILL.md} from the target context and retrieved evidence, and supplies it to a fixed solver. Meta prompt optimization uses rollout feedback to improve benchmark-specific synthesis slots through bounded atomic edits.}
    \label{fig:skilltta-overview}
\end{figure*}

\subsection{Base Test-Time Skill Synthesis}

SkillTTA synthesizes one temporary skill for each test task. Given a pool of training tasks from the same family, each record contains a description $x_i$, metadata $m_i$, a trajectory $\tau_i$, and, optionally, a success label $y_i$. For a test task $q$, the agent observes only $(x_q,m_q)$, retrieves related trajectories $R_q$, and constructs a task-specific skill without updating the solver parameters:
\[
s_q = G_{\theta}(x_q, m_q, R_q),
\]
where $G_{\theta}$ denotes the skill-synthesis model. A fixed solver conditions on $(x_q,m_q,s_q)$ to produce the final actions or answer.

Retrieval provides evidence from the same task family while avoiding target leakage. It uses only target-visible fields and excludes hidden tests, gold answers, target trajectories, and other evaluator-only information.

\subsection{Meta Prompt Optimization}

Building on base synthesis, MPO updates a benchmark-specific synthesis prompt from paired solver feedback rather than optimizing one global skill. Algorithm~\ref{alg:skilltta} summarizes MPO and test-time inference.

\begin{algorithm}[t]
\caption{SkillTTA with meta prompt optimization.}
\label{alg:skilltta}
\begin{algorithmic}[1]
\renewcommand{\algorithmicrequire}{\textbf{Input:}}
\REQUIRE pool $D$, task $q$, initial slot library $P$
\FOR{$epoch = 1,\ldots,E$ \textbf{ (MPO)}}
    \STATE sample training batch $B$ from $D$
    \FOR{each task $b \in B$}
        \STATE retrieve $R_b$ and synthesize $s_b=G(x_b,m_b,R_b;P)$
        \STATE evaluate $s_b$ with the fixed solver
    \ENDFOR
    \STATE compare current and previous paired runs
    \STATE critic attributes sampled transitions
    \STATE editor applies at most $K$ atomic slot edits to $P$
\ENDFOR
\STATE \textbf{Test-time:}
\STATE retrieve $R_q$ and synthesize $s_q=G(x_q,m_q,R_q;P)$
\STATE \textbf{return} the fixed-solver output for $(x_q,m_q,s_q)$
\end{algorithmic}
\end{algorithm}

\begin{figure}[t]
    \centering
    \includegraphics[width=0.98\columnwidth]{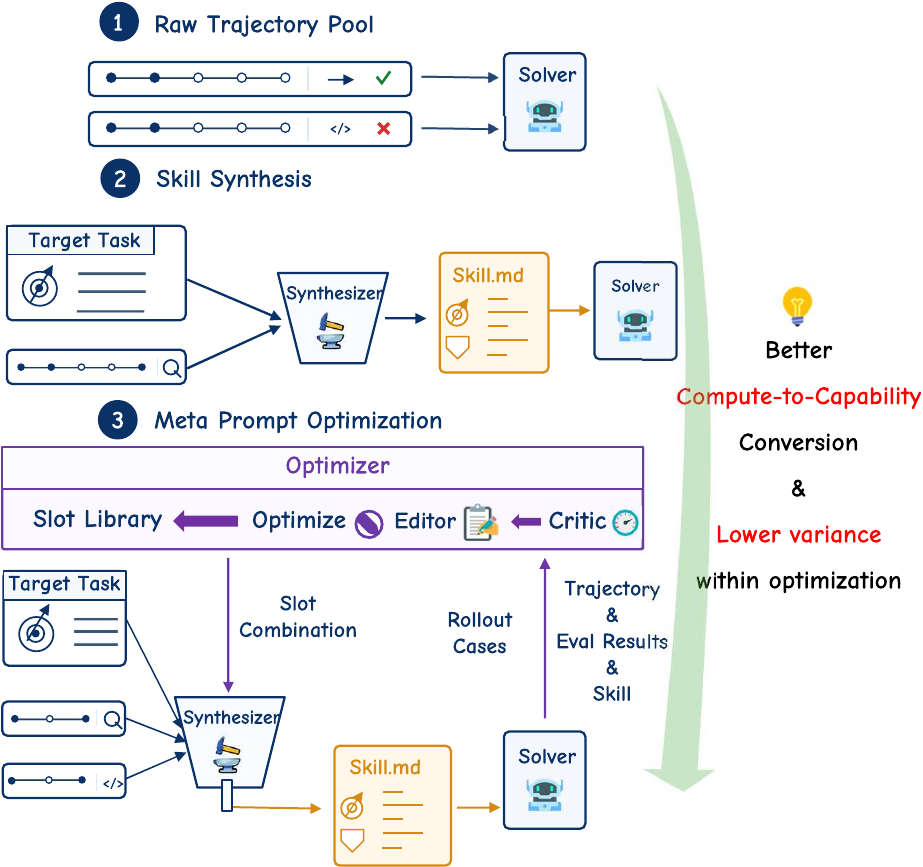}
    \caption{Three levels of compute-to-capability conversion. Raw trajectory pooling exposes evidence, skill synthesis converts evidence into task-conditioned skills, and meta prompt optimization improves the synthesis policy itself.}
    \label{fig:cost-level}
\end{figure}

\begin{figure}[t]
    \centering
    \includegraphics[width=0.88\columnwidth]{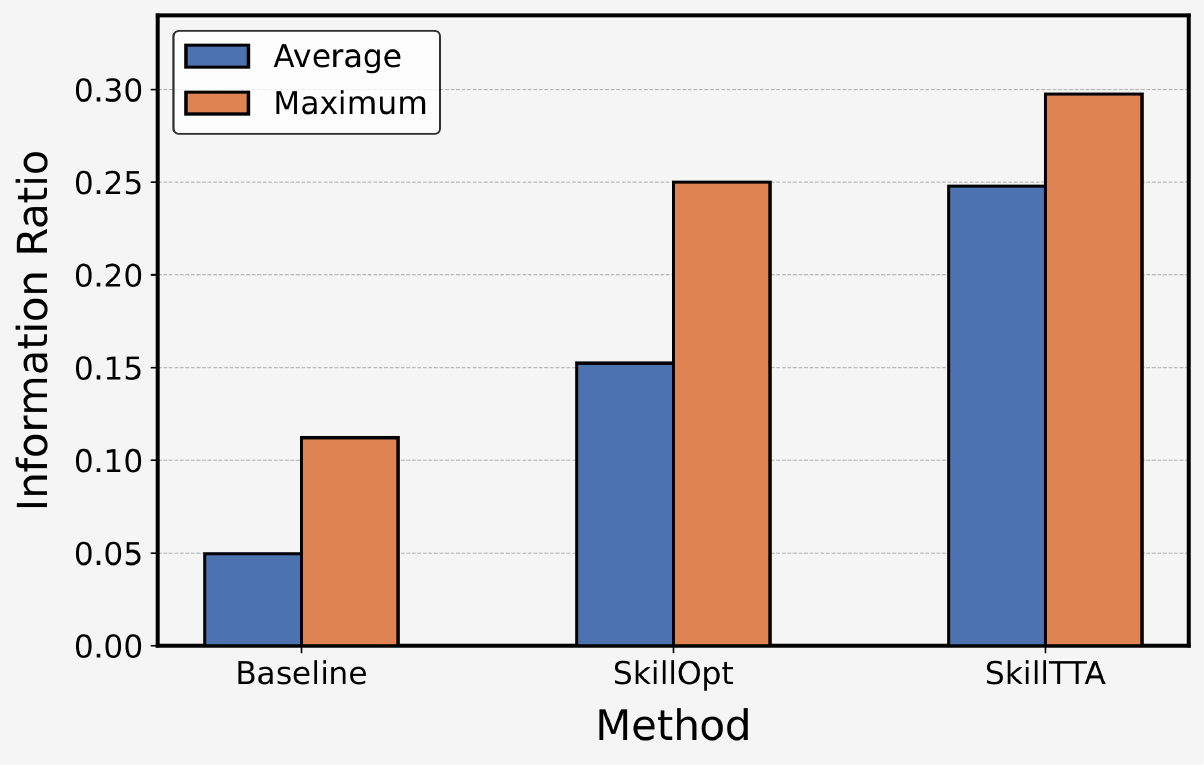}
    \caption{Optimization-signal information ratio, measured as $|\mathbb{E}[\Delta]|/\operatorname{Std}(\Delta)$ for paired task-level edit responses. Higher values indicate that an optimizer can more reliably distinguish useful edits under a finite evaluation budget.}
    \label{fig:information-ratio}
\end{figure}

Figure~\ref{fig:cost-level} places MPO within three levels of compute-to-capability conversion: experience pooling, skill synthesis, and meta prompt optimization. The first level expands and selects evidence from an experience pool, whose entries may be raw trajectories, skills, or other static reusable records. The second performs plain skill synthesis, converting selected evidence into a temporary task-conditioned skill. The third uses MPO to adapt this synthesis policy to benchmark-specific evaluator contracts, action grammars, and output formats. Moving up the hierarchy enlarges the adaptation space and can therefore raise the performance ceiling without changing the deployed solver.

Using a plain \texttt{SKILL.md} file as the synthesized object keeps the policy compatible with heterogeneous task families and solvers while preserving the solver interface.

\begin{figure*}[!t]
    \centering
    \includegraphics[width=0.90\textwidth]{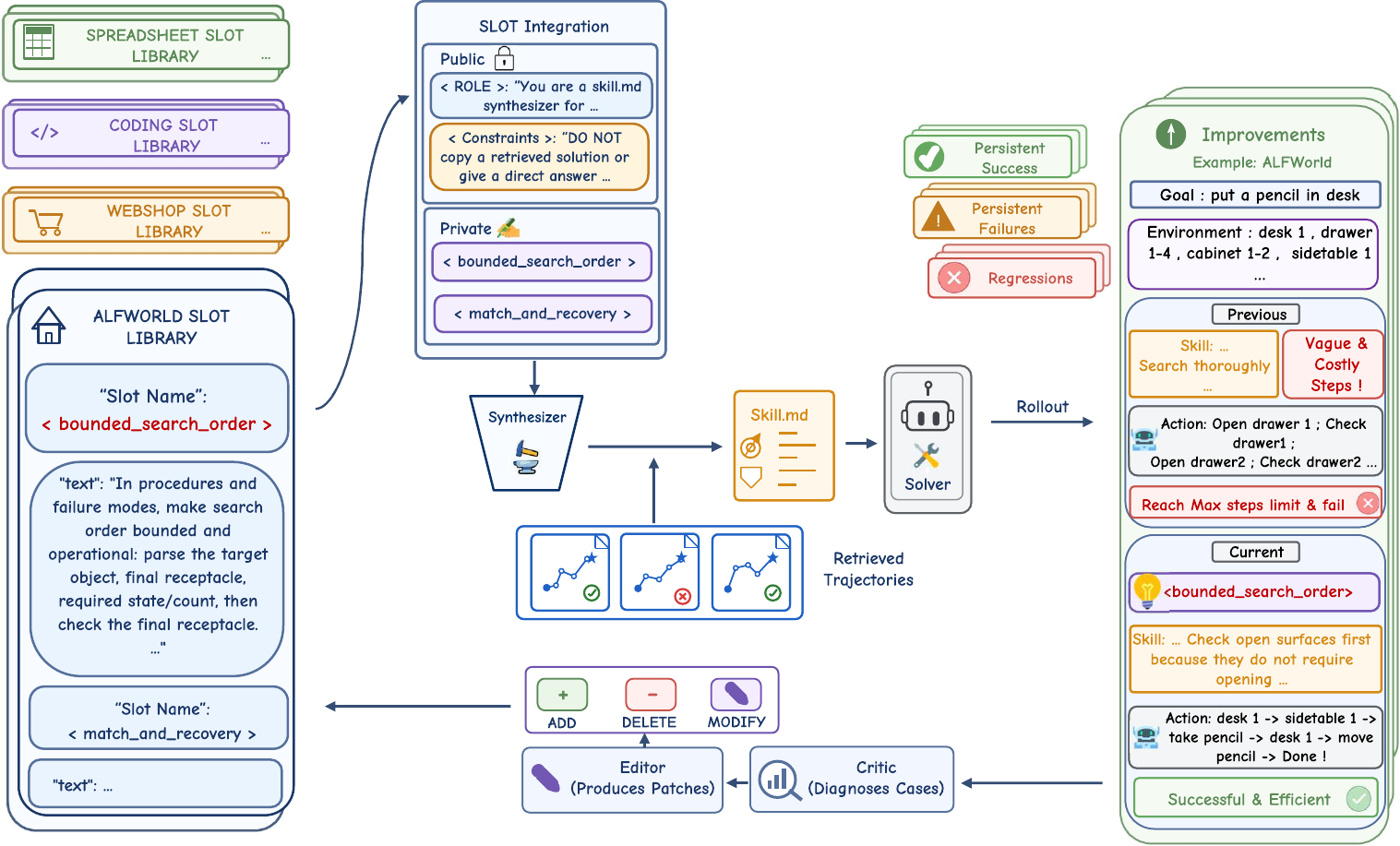}
    \caption{Meta prompt optimization updates benchmark-specific skill-synthesis slots. Each benchmark maintains a slot library, receives critic feedback, and applies bounded atomic edits to generate temporary skills.}
    \label{fig:mpo-module}
\end{figure*}

Figure~\ref{fig:mpo-module} illustrates how MPO improves the prompt used to generate $s_q$. It combines a fixed framework with a benchmark-specific slot library. The framework defines the system- and user-message skeleton, the required \texttt{SKILL.md} sections, and two slot categories: \emph{input} slots that specify how to read the target and retrieved evidence, and \emph{output} slots that specify how to write the skill.

Each benchmark updates its slot library using evaluation feedback. The optimizer compares the current and previous runs and samples cases from the observed transitions. For each case, the critic uses the paired skills, trajectories, evaluation signals, retrieved evidence, and slot diff to attribute the outcome to retrieval, the synthesized skill, solver execution, or other factors. When the evidence permits, the critic also identifies the implicated slot. An editor then converts these attributions into a bounded patch of atomic slot operations, and the resulting slot library is evaluated as a new skill-synthesis prompt. Bounded edits keep the changes auditable and limit prompt drift, especially when several budget dimensions are increased simultaneously.

The library is optimized separately for each dataset because useful edits depend on benchmark-specific requirements: ALFWorld requires adherence to its action grammar and receptacle preconditions; SpreadsheetBench requires evaluator-visible cell values and workbook-layout constraints; BigCodeBench requires compliance with function signatures and code-level contracts; and WebShop requires option matching and supporting product evidence.

\subsection{Variance Analysis and Control}

Because each edit is scored only after skill synthesis and a solver rollout, MPO expands the search space but also introduces noise during optimization. Let $Y$ denote the score of an MPO candidate, $Q$ the sampled task and retrieval context, $E$ the slot edit, and $S$ the synthesized skill. By the law of total variance,
\[
\begin{aligned}
\operatorname{Var}(Y)
=&\ \operatorname{Var}_{Q}(\mu_Q) \\
&+ \mathbb{E}_{Q}\operatorname{Var}_{E}(\mu_{Q,E}) \\
&+ \mathbb{E}_{Q,E}\operatorname{Var}_{S}(\mu_{Q,E,S}) \\
&+ \mathbb{E}_{Q,E,S}\operatorname{Var}(Y\mid Q,E,S),
\end{aligned}
\]
with $\mu_Q=\mathbb{E}[Y\mid Q]$, $\mu_{Q,E}=\mathbb{E}[Y\mid Q,E]$, and $\mu_{Q,E,S}=\mathbb{E}[Y\mid Q,E,S]$.

The additional term, relative to directly optimizing a single global skill, arises from mapping an edited meta prompt to a synthesized skill. A global-skill optimizer edits the skill document itself, so the corresponding decomposition has no intermediate variable $S$:
\[
\begin{aligned}
\operatorname{Var}(Y)
=&\ \operatorname{Var}_{Q}(\widetilde{\mu}_Q) \\
&+\mathbb{E}_{Q}\operatorname{Var}_{E}(\widetilde{\mu}_{Q,E}) \\
&+\mathbb{E}_{Q,E}\operatorname{Var}(Y\mid Q,E).
\end{aligned}
\]
SkillTTA instead evaluates the indirect path $E\rightarrow S\rightarrow Y$. It therefore introduces
\[
V_{\mathrm{syn}}
=\mathbb{E}_{Q,E}\operatorname{Var}_{S}(\mu_{Q,E,S}),
\]
which measures the rollout-relevant variation introduced when the slot-edited meta prompt is translated into a temporary skill.

The first term, the task-sampling variance $\operatorname{Var}_Q(\mu_Q)$, arises in any naive comparison of two prompts evaluated on independently sampled tasks; the remaining terms capture edit choice, skill synthesis, and downstream solver and evaluator noise. SkillTTA removes the first term by scoring every edit on \emph{paired} transitions over the same tasks. For an edit $e:p\rightarrow p'$ with benchmark-normalized utility $U_b$,
\[
\Delta_q(e)=\operatorname{sgn}_{\epsilon}\!\left(U_b(Y_q(p'))-U_b(Y_q(p))\right)\in\{-1,0,1\},
\]
where $\Delta=1$ denotes an improvement, $-1$ a regression, and $0$ no change within a tolerance $\epsilon$ set at the evaluator noise floor. The sign preserves robustness to evaluator noise while retaining the direction needed to label an edit as helpful or harmful. Tasks with $\Delta=0$ do not reveal whether the skill policy had a positive or negative effect, so the raw variance of $\Delta$ is uninformative on its own. We therefore track the information ratio
\[
\mathcal{I}(\Delta)=
\frac{|\mathbb{E}[\Delta]|}{\operatorname{Std}(\Delta)},
\]
a paired effect size that expresses the mean edit effect in units of its own noise. This signal reflects the joint behavior of the solver, skill synthesizer, and meta-prompt optimizer. Three additional controls increase the signal by reducing residual variance from editing, synthesis, and execution.

\textbf{First, per-dataset optimization} avoids cross-benchmark cancellation. Over a random benchmark $B$,
\[
\operatorname{Var}(\Delta(e))
=
\mathbb{E}_{B}\operatorname{Var}(\Delta(e)\mid B)
+
\operatorname{Var}_{B}(\mathbb{E}[\Delta(e)\mid B]).
\]
Pooling an edit across benchmarks adds the between-benchmark term $\operatorname{Var}_{B}(\mathbb{E}[\Delta(e)\mid B])\ge0$ to the denominator. When benchmarks disagree in sign, the pooled mean can also shrink toward zero, reducing the numerator. Conditioning on $B=b$ removes both effects, so each update measures the effect of the edit on a single benchmark.

\textbf{Second, oversampling nonzero transitions} raises the signal directly. With $p_+=P(\Delta=1)$, $p_-=P(\Delta=-1)$, $\rho=p_++p_-$, and $\mu=p_+-p_-$, we have $\mathbb{E}[\Delta]=\mu$, $\mathbb{E}[\Delta^2]=\rho$, so
\[
\mathcal{I}(\Delta)=\frac{|\mu|}{\sqrt{\rho-\mu^2}}.
\]
Resampling improvements and regressions by a common factor $\alpha\in[1,1/\rho]$ preserves their relative balance and ensures that $p_0=1-\alpha\rho\ge0$, giving $\mu_{\alpha}=\alpha\mu$, $\rho_{\alpha}=\alpha\rho$, and
\[
\mathcal{I}_{\alpha}^{2}
=
\frac{\alpha\mu^2}{\rho-\alpha\mu^2},
\qquad
\frac{\partial \mathcal{I}_{\alpha}^{2}}{\partial \alpha}
=
\frac{\mu^2\rho}{(\rho-\alpha\mu^2)^2}>0 .
\]
The signal therefore increases monotonically up to $\alpha=1/\rho$, at which point unchanged tasks disappear. The critic consequently operates on a transition-enriched sample rather than the raw run.

\textbf{Third, bounded atomic edits} keep credit assignment identifiable. Under a local effect model, the paired transition decomposes as
\[
\Delta_q(A)
=
\sum_{j\in A}\tau_{j,q}
+
\sum_{\substack{i<j\\ i,j\in A}}\tau_{ij,q}
+
\eta_q.
\]
Thus, an update of size $|A|\le K$ exposes $O(K^2)$ unknown effects, compared with $O(d^2)$ for the full $d$-slot library. Here, $K$ is the maximum number of slot edits in a single optimizer patch. Under a fixed transition budget, the estimation variance for each effect grows with the number of competing terms, so capping $K\ll d$ improves attribution.

Together, these controls make the higher-capacity synthesis policy easier to optimize under a finite evaluation budget. Figure~\ref{fig:information-ratio} shows that SkillTTA achieves higher mean and peak information ratios than SkillOpt and the baseline. SkillOpt directly optimizes one global skill, whereas SkillTTA uses MPO to optimize a separate skill-synthesis policy for each benchmark. Complementary ablations in Table~\ref{tab:skill-ablation} isolate the effects of bounded atomic slots, oversampled transition signals, and per-dataset optimization.

\section{Experiments}
\label{sec:experiments}

\subsection{Experimental Setup}

\paragraph{Datasets and metrics.}
The evaluation covers four task families: SpreadsheetBench for spreadsheet manipulation \citep{spreadsheetbench2024}, ALFWorld for text-based household interaction \citep{alfworld2021}, BigCodeBench for practical Python generation \citep{bigcodebench2024}, and WebShop for grounded product search \citep{webshop2022}. We report Pass@1 for SpreadsheetBench and BigCodeBench, success rate and the number of steps in successful episodes for ALFWorld, and success rate and average score for WebShop.

\paragraph{Baselines.}
Zero-shot and ReAct use the fixed solver directly; RAG retrieves evidence without skill synthesis; TARSE retrieves skills and experiences for test-time reasoning; Trace2Skill and ExpeL provide static synthesized skills or experience-derived insights; SkillX and SkillOpt optimize a global reusable skill rather than synthesizing a temporary per-task skill; and MemRL learns memory utilities through interaction. Together, these baselines cover raw retrieval, retrieved skill reuse, static skill reuse, global skill optimization, and runtime memory learning.

\paragraph{Implementation details.}
Unless stated otherwise, the task solver is GPT-5.4-mini, retrieval uses top-$k=3$, and the skill-synthesis model is one of GPT-5.4-mini, DeepSeek-V4-Pro, and GPT-5.5. In the cost plots, every LLM role---the solver, skill synthesizer, MPO critic, and MPO editor---uses DeepSeek-V4-Flash, and output tokens are priced at twice the rate of input tokens. This fixed-backbone comparison rules out an explanation in which a stronger teacher model directly solves the task and passes answer content to the solver.

\begin{table*}[!t]
    \centering
    {\small
    \setlength{\tabcolsep}{2.6pt}
    \begin{tabularx}{\textwidth}{l*{6}{>{\centering\arraybackslash}X}}
        \toprule
        \textbf{Method} &
        \multicolumn{2}{c}{\textbf{ALFWorld}} &
        \textbf{Spreadsheet} &
        \textbf{BigCodeBench} &
        \multicolumn{2}{c}{\textbf{WebShop}} \\
        \cmidrule(lr){2-3}\cmidrule(lr){4-4}\cmidrule(lr){5-5}\cmidrule(l){6-7}
        & \textbf{Succ. $\uparrow$} & \textbf{Steps $\downarrow$} & \textbf{Pass@1 $\uparrow$} & \textbf{Pass@1 $\uparrow$} & \textbf{Succ. $\uparrow$} & \textbf{Score $\uparrow$} \\
        \midrule
        Zero-shot & 0.643 & 11.02 & 0.363 & 0.489 & 0.292 & 0.585 \\
        ReAct & 0.686 & 12.23 & 0.418 & 0.503 & 0.272 & 0.563 \\
        MemRL & 0.907 & 11.85 & 0.387 & 0.413 & 0.334 & 0.597 \\
        RAG & 0.836 & 10.20 & 0.422 & 0.497 & 0.324 & 0.599 \\
        TARSE & 0.793 & 9.20 & 0.440 & 0.523 & 0.332 & 0.595 \\
        \midrule
        \multicolumn{7}{c}{\emph{Synthesis Model: GPT-5.4-mini}} \\
        Trace2Skill & 0.693 & 13.59 & 0.420 & 0.535 & 0.320 & 0.558 \\
        ExpeL & 0.757 & 11.21 & 0.440 & 0.526 & 0.328 & 0.568 \\
        SkillX & 0.729 & 12.80 & 0.435 & 0.483 & 0.236 & 0.449 \\
        SkillOpt & 0.779 & 10.20 & 0.432 & 0.531 & 0.284 & 0.527 \\
        \rowcolor{skillttarow}
        SkillTTA & \textbf{\underline{0.814}} & \textbf{\underline{9.80}} & \textbf{\underline{0.455}} & \textbf{\underline{0.547}} & \textbf{\underline{0.356}} & \textbf{\underline{0.610}} \\
        \midrule
        \multicolumn{7}{c}{\emph{Synthesis Model: DeepSeek-V4-Pro}} \\
        Trace2Skill & 0.821 & 9.80 & 0.445 & 0.520 & 0.276 & 0.443 \\
        ExpeL & 0.721 & 9.56 & 0.420 & 0.503 & 0.334 & 0.618 \\
        SkillX & 0.714 & 13.30 & 0.425 & 0.501 & 0.352 & 0.625 \\
        SkillOpt & 0.821 & 9.72 & 0.455 & 0.519 & 0.304 & 0.509 \\
        \rowcolor{skillttarow}
        SkillTTA & \textbf{\underline{0.921}} & \textbf{\underline{8.80}} & \textbf{\underline{0.550}} & \textbf{\underline{0.592}} & \textbf{\underline{0.369}} & \textbf{\underline{0.643}} \\
        \midrule
        \multicolumn{7}{c}{\emph{Synthesis Model: GPT-5.5}} \\
        Trace2Skill & 0.713 & 12.74 & 0.397 & 0.517 & 0.336 & 0.562 \\
        ExpeL & 0.812 & 8.60 & 0.445 & 0.515 & 0.324 & 0.532 \\
        SkillX & 0.750 & 12.00 & 0.395 & 0.498 & 0.324 & 0.538 \\
        SkillOpt & 0.858 & 9.17 & 0.475 & 0.540 & 0.280 & 0.507 \\
        \rowcolor{skillttarow}
        SkillTTA & \textbf{\underline{0.936}} & \textbf{\underline{8.30}} & \textbf{\underline{0.610}} & \textbf{\underline{0.645}} & \textbf{\underline{0.384}} & \textbf{\underline{0.647}} \\
        \bottomrule
    \end{tabularx}
    }
    \caption{Main benchmark results. ALFWorld reports success and steps; SpreadsheetBench and BigCodeBench report Pass@1; and WebShop reports success and score. }
    \label{tab:main-results}
\end{table*}

\subsection{Main Results}

Table~\ref{tab:main-results} summarizes the main comparison. The GPT-5.5 SkillTTA configuration achieves the highest task scores among the methods listed, reaching a success rate of 0.936 on ALFWorld, Pass@1 scores of 0.610 on SpreadsheetBench and 0.645 on BigCodeBench, and a success rate of 0.384 with an average score of 0.647 on WebShop. Within each skill-synthesis block, SkillTTA improves on Trace2Skill, ExpeL, SkillX, and SkillOpt across every metric, supporting the value of conditioning the skill on the current task. It also outperforms the fixed baselines, which do not require a separate synthesis model.

\begin{figure*}[!t]
    \centering
    \includegraphics[width=0.94\textwidth]{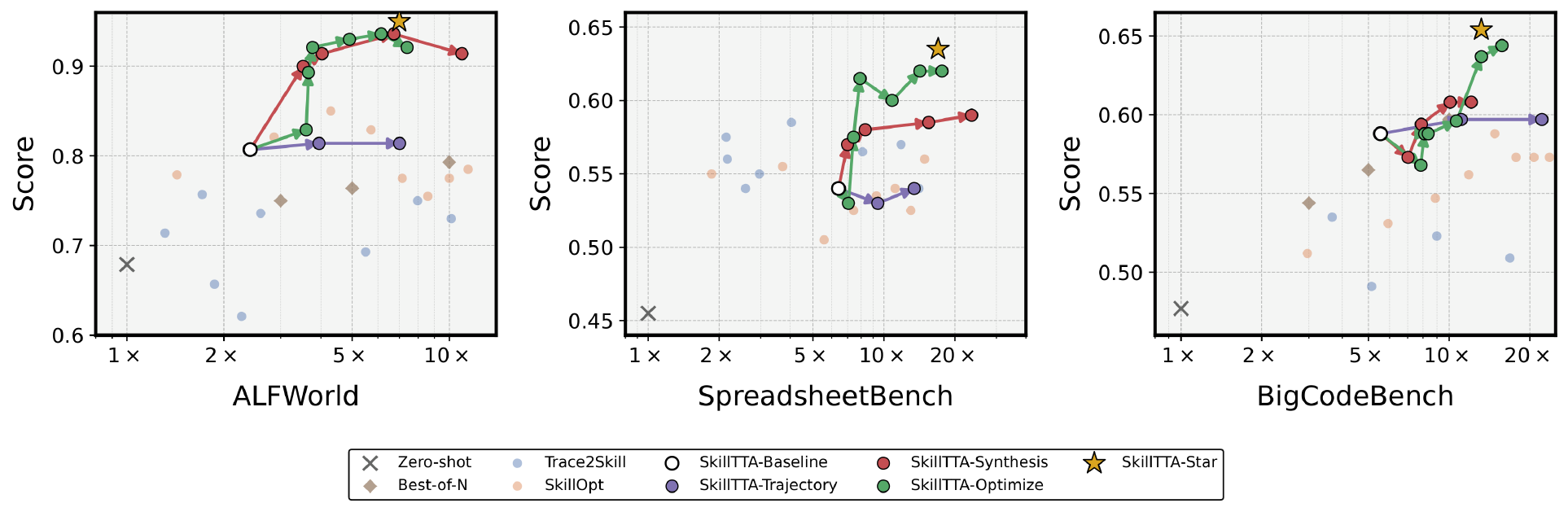}
    \caption{Cost--score tradeoff on ALFWorld, SpreadsheetBench, and BigCodeBench, with DeepSeek-V4-Flash used for the solver, skill synthesizer, MPO critic, and MPO editor. The $x$-axis shows token cost relative to zero-shot inference. Faded markers denote baseline reuse and sampling strategies; the bold path traces increases in the trajectory-pool, skill-synthesis, and optimization budgets up to SkillTTA-Star.}
    \label{fig:cost-tradeoff}
\end{figure*}

\begin{table*}[t]
    \centering
    {\small
    \setlength{\tabcolsep}{3.4pt}
    \begin{tabular*}{\textwidth}{@{\extracolsep{\fill}}lcccccc@{}}
        \toprule
        \textbf{Variant} &
        \multicolumn{2}{c}{\textbf{ALFWorld}} &
        \textbf{Spreadsheet} &
        \textbf{BigCodeBench} &
        \multicolumn{2}{c}{\textbf{WebShop}} \\
        \cmidrule(lr){2-3}\cmidrule(lr){4-4}\cmidrule(lr){5-5}\cmidrule(l){6-7}
        & \textbf{Succ. $\uparrow$} & \textbf{Steps $\downarrow$} & \textbf{Pass@1 $\uparrow$} & \textbf{Pass@1 $\uparrow$} & \textbf{Succ. $\uparrow$} & \textbf{Score $\uparrow$} \\
        \midrule
        Raw trajectories (RAG) & 0.836 & 10.20 & 0.422 & 0.497 & 0.324 & 0.599 \\
        SkillTTA (non-MPO) & 0.850 & 9.90 & 0.545 & 0.623 & 0.340 & 0.572 \\
        SkillTTA w/o retrieval (test task only) & 0.795 & 12.90 & 0.460 & 0.591 & 0.336 & 0.581 \\
        SkillTTA w/ random retrieval & 0.909 & 8.60 & 0.560 & 0.602 & 0.350 & 0.624 \\
        MPO w/o atomic slots & 0.871 & 8.88 & 0.505 & 0.614 & 0.352 & 0.603 \\
        MPO w/o oversampled signals & 0.886 & 9.60 & 0.540 & 0.632 & 0.364 & 0.619 \\
        MPO w/o per-dataset opt. & 0.871 & 9.10 & 0.575 & 0.611 & 0.368 & 0.606 \\
        SkillTTA (MPO) & \textbf{\underline{0.936}} & \textbf{\underline{8.30}} & \textbf{\underline{0.610}} & \textbf{\underline{0.645}} & \textbf{\underline{0.384}} & \textbf{\underline{0.647}} \\
        \bottomrule
    \end{tabular*}
    }
    \caption{Ablation of skill synthesis, retrieval, and prompt quality. The retrieval rows retain the SkillTTA synthesis and MPO settings and vary only the retrieval policy. BigCodeBench reports Pass@1, and WebShop reports success and score.}
    \label{tab:skill-ablation}
\end{table*}

\begin{table}[t]
    \centering
    {\small
    \setlength{\tabcolsep}{4pt}
    \begin{tabular}{lcccc}
        \toprule
        \textbf{Top-$k$} & \textbf{ALF Succ. $\uparrow$} & \textbf{Steps $\downarrow$} & \shortstack{\textbf{Sheet}\\\textbf{Pass@1 $\uparrow$}} & \shortstack{\textbf{Shop}\\\textbf{Succ./Score $\uparrow$}} \\
        \midrule
        1 & 0.929 & 8.50 & 0.640 & 0.360/0.614 \\
        3 & 0.936 & \textbf{\underline{8.30}} & 0.610 & 0.384/\textbf{\underline{0.647}} \\
        5 & 0.936 & \textbf{\underline{8.30}} & 0.635 & 0.380/0.627 \\
        9 & \textbf{\underline{0.943}} & \textbf{\underline{8.30}} & \textbf{\underline{0.645}} & \textbf{\underline{0.396}}/0.636 \\
        \bottomrule
    \end{tabular}
    }
    \caption{Retrieval depth ablation. WebShop reports success and score.}
    \label{tab:topk}
\end{table}

\subsection{Compute-to-Capability Conversion}

Figure~\ref{fig:cost-tradeoff} compares cost and score on ALFWorld, SpreadsheetBench, and BigCodeBench. Whereas Table~\ref{tab:main-results} uses three skill-synthesis models to test compatibility across model capabilities and providers, this analysis uses DeepSeek-V4-Flash for every LLM role: the task solver, the skill synthesizer, and the MPO components responsible for critique and editing. The standard trajectory pool contains the same number of tasks as the test set, allowing the optimization, skill-synthesis, and pool-construction costs to be meaningfully amortized over the evaluated tasks. This setup standardizes cost accounting, removes model choice as a confounder, and rules out a direct-teacher explanation in which the gains come from a stronger model leaking answers to the solver. The supplementary appendix reports the full input and output token counts for all plotted points. Best-of-$N$ benefits from repeated solver sampling on ALFWorld and BigCodeBench but remains below the SkillTTA frontier at comparable or higher costs. SkillOpt and Trace2Skill also trail SkillTTA under matched cost budgets, and their cost frontiers saturate at lower capability levels. Along the conversion path, increasing the trajectory-pool or prompt size yields only small or inconsistent gains. Increasing the trajectory-pool, skill-synthesis, and optimization budgets along the three SkillTTA axes raises the frontier to 0.936/0.620/0.644. SkillTTA-Star combines these axes and reaches 0.950/0.635/0.654 with a lower convergence cost than the baseline reuse or sampling strategies. Thus, additional compute is most useful when it increases agent capability through task-conditioned skill synthesis and synthesis-policy optimization, especially when all three SkillTTA axes receive larger budgets, rather than when it only expands the trajectory pool or adds optimization epochs.

\subsection{Mechanism Ablations}

Table~\ref{tab:skill-ablation} isolates the mechanism across all benchmarks. Raw trajectories underperform SkillTTA (MPO), supporting the shift from exposing evidence to synthesizing skills. The non-MPO row removes the optimized synthesis policy, showing that synthesis requires an optimized policy rather than retrieved evidence alone. SkillTTA with random retrieval also underperforms SkillTTA with embedding-based retrieval, indicating that retrieval quality still matters after synthesis; however, it remains substantially above SkillTTA without retrieval, showing that retrieved trajectories provide valuable information, while the current embedding-only retrieval method still leaves room for better retrieval policy design. Removing atomic slot management, oversampled regression and improvement signals, or per-dataset optimization reduces the final scores. These results support the role of each component in keeping MPO edits auditable, scoring them reliably, and matching them to benchmark-specific failure modes under a limited evaluation budget. The per-dataset ablation further suggests that optimizing separately for each dataset is more effective than seeking a single policy that generalizes across all datasets.

Table~\ref{tab:topk} reports retrieval-depth sensitivity. ALFWorld and SpreadsheetBench benefit from larger $k$, while WebShop reaches its best score at top-3 and its best success at top-9. The main comparison nevertheless uses top-3 for SkillTTA to match the default retrieval budget of the other retrieval-based baselines.

\section{Conclusion}

SkillTTA treats test-time experience not as context to be replayed verbatim, but as evidence from which to construct task-specific agent capability. It retrieves related trajectories, synthesizes a temporary skill after observing the target task, and uses benchmark-specific MPO to improve the policy responsible for this conversion while leaving the solver unchanged. Because the synthesized skill is task scoped, it can be discarded after execution without permanently altering the solver or accumulating task-specific content in shared memory. Across four agent benchmarks and three synthesis-model backbones, this design consistently improves over raw retrieval, static skills, and globally optimized skills. The ablations further link these gains to the information in retrieved trajectories, per-dataset specialization, transition-enriched feedback, and bounded atomic edits. The cost analysis also shows that additional compute is more effective when invested jointly in evidence, synthesis, and synthesis-policy optimization. These results point to a practical principle for test-time adaptation: use additional compute to improve the conversion from experience to task-specific agent capability, rather than merely retrieving more experience or compressing it with a more complex procedure.

\fi

\ifincludeappendix
\appendix

\section{Appendix}

\subsubsection{Initial Meta Prompt Setup.}

This appendix records the initial prompt configuration, optimizer prompt templates, and dataset defaults. MPO starts with a fixed framework and a minimal general-purpose slot library. The framework remains unchanged and defines the prompt skeleton, output sections, slot categories, and slot budgets. MPO edits only the ordered slots in the benchmark-specific library. Each update changes at most four slots; the input and output categories each contain at most eight slots, with each slot limited to 1,200 characters.

\subsubsection{Optimization Prompt Templates.}

During MPO, the optimizer uses two prompt templates from \url{prompt_optimizer/optimizer_prompts.py}. The case critic receives a sampled task transition and returns an attribution in JSON format. The slot editor receives the critic outputs, run summary, current slot library, and most recent slot diff, and then returns a bounded JSON patch. Each user prompt below is followed by a serialized \texttt{Payload}.

\subsubsection{Splits and Defaults.}

SpreadsheetBench uses Verified-400, with 200 training and 200 test tasks. ALFWorld samples 1,000 training trajectories from the original training set using seed 42 and evaluates on valid-seen. BigCodeBench uses instruct-full, with 798 training and 342 test tasks sampled using seed 123. WebShop task generation uses seed 0; its 250 training and 250 test tasks are split from the original 500 test tasks using seed 42. Whenever possible, these choices follow the default splitting and partitioning protocols of the corresponding baselines, including the Trace2Skill split for SpreadsheetBench and the MemRL split for BigCodeBench. All trajectory pools, MPO updates, critic-visible ground truth, retrieved examples, and prompt-optimization feedback use only training tasks; test tasks remain fully held out until the final evaluation. All baselines use the same trajectory format and a pool-to-test-task ratio of 1:1. The shared format includes the task statement, input trajectory, and any ground truth provided by the benchmark. Unless an ablation changes this setting, all retrieval-based methods, including SkillX, RAG, and MemRL, use top-$k=3$ retrieval. Retrieval uses only visible task fields: SpreadsheetBench uses the instruction, instruction type, and output position; ALFWorld uses the goal object, receptacle, room, goal sentence, and subgoals; BigCodeBench uses the visible instruction, code prompt, and library list while excluding tests and canonical solutions; and WebShop uses the shopping instruction and visible product-search context. Multi-epoch methods, including MemRL, SkillOpt, and SkillTTA, use three epochs in the main table. By default, each MPO epoch samples 25\% of the training set to form the update signal. We conduct multiple runs with different random seeds and report the averaged results. The temperature is 0 for the solver and skill synthesizer and 0.2 for the MPO critic and optimizer. The maximum number of steps is 30 for ALFWorld, except for the limit of 15 used in Figure~\ref{fig:cost-tradeoff}, and 15 for SpreadsheetBench and WebShop. MPO critic cases are sampled at a 3:2:1:1 ratio across regressions, improvements, persistent successes, and persistent failures.

\subsubsection{Figure 6 Token Accounting.}

Table~\ref{tab:fig5-token-accounting} reports the detailed data for the points in Figure~\ref{fig:cost-tradeoff}. Let $I$ and $O$ denote the input and output token totals for a given point. Its relative cost is $(I+2O)/(I_0+2O_0)$, where $(I_0,O_0)$ denotes the corresponding zero-shot denominator. T2S denotes Trace2Skill, SOpt denotes SkillOpt, and TTA denotes SkillTTA.

\begin{table*}[t]
    \centering
    \scriptsize
    \setlength{\tabcolsep}{1.45pt}
    \begin{tabular*}{\textwidth}{@{\extracolsep{\fill}}lrrrrrrrrrrrr@{}}
        \toprule
        & \multicolumn{4}{c}{\textbf{ALFWorld}} &
        \multicolumn{4}{c}{\textbf{SpreadsheetBench}} &
        \multicolumn{4}{c}{\textbf{BigCodeBench}} \\
        \cmidrule(lr){2-5}\cmidrule(lr){6-9}\cmidrule(l){10-13}
        \textbf{Point} &
        \textbf{Score} & \textbf{In} & \textbf{Out} & \textbf{Rel.} &
        \textbf{Score} & \textbf{In} & \textbf{Out} & \textbf{Rel.} &
        \textbf{Score} & \textbf{In} & \textbf{Out} & \textbf{Rel.} \\
        \midrule
        \multicolumn{13}{l}{\emph{Background reuse and sampling baselines}} \\
        Zero-shot & 0.679 & 13515 & 341 & 1.000 & 0.455 & 4901 & 1437 & 1.000 & 0.477 & 1651 & 364 & 1.000 \\
        Best-of-3 & 0.750 & 40545 & 1023 & 3.000 & -- & -- & -- & -- & 0.544 & 4953 & 1092 & 3.000 \\
        Best-of-5 & 0.764 & 67575 & 1705 & 5.000 & -- & -- & -- & -- & 0.565 & 8255 & 1820 & 5.000 \\
        Best-of-10 & 0.793 & 135150 & 3410 & 10.000 & -- & -- & -- & -- & 0.597 & 16510 & 3640 & 10.000 \\
        T2S-0.5max & 0.714 & 9798.5 & 4419.5 & 1.313 & 0.560 & 5559.5 & 5637.5 & 2.165 & -- & -- & -- & -- \\
        T2S-0.75max & 0.621 & 15856.2 & 8172.8 & 2.268 & 0.550 & 8004.8 & 7498.8 & 2.958 & -- & -- & -- & -- \\
        T2S-1max & 0.736 & 19685 & 8613 & 2.600 & 0.585 & 11123 & 10163 & 4.045 & 0.523 & 5334 & 8038 & 9.000 \\
        T2S-1high & 0.657 & 19731.3 & 3407 & 1.870 & 0.540 & 10888 & 4613 & 2.587 & 0.491 & 5334 & 3445 & 5.138 \\
        T2S-1no & 0.757 & 19731.3 & 2286 & 1.712 & 0.575 & 10888 & 2879 & 2.141 & 0.535 & 5334 & 1688 & 3.661 \\
        T2S-2max & 0.693 & 40322 & 18895 & 5.502 & 0.565 & 21361 & 20823 & 8.104 & 0.509 & 9879 & 15121 & 16.865 \\
        T2S-3max & 0.750 & 60105 & 26578 & 7.978 & 0.570 & 29621 & 31072 & 11.803 & -- & -- & -- & -- \\
        T2S-3b4 & 0.730 & 62654 & 40740 & 10.152 & 0.540 & 30865 & 39208 & 14.055 & -- & -- & -- & -- \\
        SOpt-1 & 0.779 & 14774 & 2765 & 1.430 & 0.550 & 7687 & 3372 & 1.856 & 0.512 & 1732.8 & 2656 & 2.961 \\
        SOpt-2 & 0.821 & 29548 & 5530 & 2.860 & 0.555 & 15374 & 6744 & 3.712 & 0.531 & 3465.5 & 5312 & 5.922 \\
        SOpt-3 & 0.850 & 44322 & 8295 & 4.290 & 0.505 & 23061 & 10116 & 5.568 & 0.547 & 5198.2 & 7968 & 8.884 \\
        SOpt-4 & 0.829 & 59096 & 11060 & 5.721 & 0.525 & 30748 & 13488 & 7.424 & 0.562 & 6931 & 10624 & 11.845 \\
        SOpt-5 & 0.775 & 73870 & 13825 & 7.151 & 0.535 & 38435 & 16860 & 9.280 & 0.588 & 8663.8 & 13280 & 14.806 \\
        SOpt-6 & 0.755 & 88644 & 16590 & 8.581 & 0.540 & 46122 & 20232 & 11.136 & 0.573 & 10396.5 & 15936 & 17.767 \\
        SOpt-7 & 0.775 & 103418 & 19355 & 10.011 & 0.525 & 53809 & 23604 & 12.993 & 0.573 & 12129.2 & 18592 & 20.729 \\
        SOpt-8 & 0.785 & 118192 & 22120 & 11.441 & 0.560 & 61496 & 26976 & 14.849 & 0.573 & 13862 & 21248 & 23.690 \\
        \midrule
        \multicolumn{13}{l}{\emph{SkillTTA conversion path}} \\
        TTA-Baseline & 0.807 & 29617 & 1881.6 & 2.412 & 0.540 & 39302 & 5278 & 6.413 & 0.588 & 8422 & 2389 & 5.549 \\
        TTA-Traj. 1 & 0.814 & 50318 & 2155.2 & 3.947 & 0.530 & 54005 & 9589 & 9.413 & 0.597 & 16844 & 4778 & 11.097 \\
        TTA-Traj. 2 & 0.814 & 91720 & 2702.4 & 7.018 & 0.540 & 73609 & 15337 & 13.413 & 0.597 & 33688 & 9556 & 22.194 \\
        TTA-Synth. 1 & 0.914 & 44787.2 & 5499.5 & 4.031 & 0.580 & 45232.5 & 9725.5 & 8.319 & 0.594 & 13058.6 & 2840.2 & 7.877 \\
        TTA-Synth. 2 & 0.900 & 44787.2 & 1964.5 & 3.520 & 0.570 & 45232.5 & 4621.8 & 7.007 & 0.573 & 13058.6 & 1839.8 & 7.036 \\
        TTA-Synth. 3 & 0.936 & 86189.2 & 3474.2 & 6.730 & 0.585 & 92676.5 & 13805.5 & 15.471 & 0.608 & 16360.6 & 3831.8 & 10.098 \\
        TTA-Synth. 4 & 0.914 & 143117 & 4226.6 & 10.952 & 0.590 & 140120.5 & 21385.5 & 23.523 & 0.608 & 19662.6 & 4559.8 & 12.098 \\
        TTA-Opt. 1 & 0.893 & 44787.2 & 2927 & 3.659 & 0.575 & 45232.5 & 6225.5 & 7.419 & 0.588 & 13058.6 & 3103.8 & 8.098 \\
        TTA-Opt. 2 & 0.930 & 59957.5 & 3972.4 & 4.907 & 0.600 & 66743 & 8661 & 10.812 & 0.596 & 17695.2 & 3818.6 & 10.648 \\
        TTA-Opt. 3 & 0.936 & 75127.8 & 5017.8 & 6.154 & 0.620 & 88253.5 & 11096.5 & 14.205 & 0.637 & 22331.8 & 4533.4 & 13.198 \\
        TTA-Opt. 4 & 0.921 & 90298 & 6063.2 & 7.401 & 0.620 & 109764 & 13532 & 17.598 & 0.644 & 26968.4 & 5248.2 & 15.748 \\
        TTA-Opt. 5 & 0.921 & 44787.2 & 3665 & 3.766 & 0.615 & 45232.5 & 8137.5 & 7.911 & 0.588 & 13058.6 & 3418.8 & 8.363 \\
        TTA-Opt. 6 & 0.829 & 44787.2 & 2522 & 3.601 & 0.530 & 45232.5 & 4862.5 & 7.068 & 0.568 & 13058.6 & 2800.8 & 7.844 \\
        TTA-Star & 0.950 & 75127.8 & 10833.3 & 6.994 & 0.635 & 88253.5 & 21732.5 & 16.941 & 0.654 & 21552.5 & 4912.8 & 13.190 \\
        \bottomrule
    \end{tabular*}
    \caption{Input and output token totals for the points rendered in Figure~\ref{fig:cost-tradeoff}. Best-of-$N$ uses $N$ zero-shot solver calls. The relative-cost column matches the Figure~\ref{fig:cost-tradeoff} axis computation, with output tokens weighted by a factor of 2.}
    \label{tab:fig5-token-accounting}
\end{table*}

\subsubsection{Prompt Listings.}

The listings below show the initial slot library, framework schema, and prompt templates.

\begin{lstlisting}[style=yamlconfig,caption={Initial general slot library from general.yaml.}]
slot_library_version: general-slots-v3
benchmark: general

dataset_identity: >-
  This benchmark provides one target task context and optional retrieved examples
  from related tasks. Synthesize a compact, reusable SKILL.md that helps a future
  agent solve the target task from its own context, not an answer to copy.

slots:
  input:
    - id: binding_contract
      text: >-
        Treat the target task context as the only binding contract for the new
        skill. Use retrieved examples only as evidence for transferable tactics
        and common failure modes; never treat them as extra requirements and
        never copy their concrete facts, ids, paths, names, or numeric values.
  output:
    - id: reusable_guidance
      text: >-
        Write each section as reusable guidance grounded in the target contract
        and retrieved evidence, so it transfers to the target task. It must not
        restate this specific task, copy a retrieved solution, or give the final
        answer.
\end{lstlisting}

\begin{lstlisting}[style=yamlconfig,caption={Optimized benchmark-specific slot libraries after MPO.}]
alfworld-slots-v5-3:
  input:
    - id: binding_contract
      text: >-
        Treat the target task context as the only binding contract for the new
        skill. Use retrieved examples only as evidence for transferable tactics
        and common failure modes; never treat them as extra requirements and
        never copy their concrete facts, ids, paths, names, or numeric values.
  output:
    - id: reusable_guidance
      text: >-
        In the Possible Procedures section, provide a step-by-step search
        strategy. First, list every receptacle type from the initial room
        observation, including open surfaces, closed containers, and uncommon
        receptacles such as toilets, garbage cans, towel holders, coffee tables,
        shelves, sidetables, and sofas. Prioritize checking open surfaces before
        closed containers; search each open surface exactly once and do not
        revisit. If the object is not found after checking a few common open
        surfaces, proceed to closed containers to avoid exhausting step limits.
        Check closed containers systematically: cabinets, then drawers, then
        fridge, then other closed containers like dressers. For multi-object
        tasks, pick up one object at a time, deliver it to the target
        immediately, then return for the next. If a move action results in
        "Nothing happened", try opening the target receptacle before retrying.
        After placing the final object, do not output any action; the
        environment will end the episode automatically.
    - id: multi_object_handling
      text: >-
        For tasks that require collecting multiple identical objects, explicitly
        instruct the agent to carry only one object at a time. After picking up
        the first object, immediately move to the target receptacle, place it,
        then return to fetch the second object. Warn that attempting to pick up
        a second object while holding one will result in "Nothing happened" and
        waste steps.

bigcodebench-slots-v2-2:
  input:
    - id: binding_contract
      text: >-
        Treat the target task context as the only binding contract for the new
        skill. Use retrieved examples only as evidence for transferable tactics
        and common failure modes; never treat them as extra requirements and
        never copy their concrete facts, ids, paths, names, or numeric values.
  output:
    - id: reusable_guidance
      text: >-
        Write each section as reusable guidance that transfers to the target
        task. Derive concrete operational steps from the target contract and
        retrieved evidence, but avoid hardcoding specific values, such as regex
        patterns, plot methods, API endpoints, or default parameters, unless the
        task requires them exactly. Use conditional phrasing to keep guidance
        flexible. Do not add edge-case handling, such as empty input checks or
        directory existence checks, unless the task specification or tests
        explicitly require it. Focus on common pitfalls and verification checks
        that are grounded in the task's test expectations.

spreadsheet-slots-v3-1:
  input:
    - id: binding_contract
      text: >-
        Treat the target task context as the only binding contract for the new
        skill. Use retrieved examples only as evidence for transferable tactics
        and common failure modes; never treat them as extra requirements and
        never copy their concrete facts, ids, paths, names, or numeric values.
  output:
    - id: reusable_guidance
      text: >-
        Write each section as reusable guidance grounded in the target contract
        and retrieved evidence, so it transfers to the target task. It must not
        restate this specific task, copy a retrieved solution, or give the final
        answer.
    - id: worksheet_analysis_guidance
      text: >-
        In the Possible procedures section, always include steps to thoroughly
        analyze the target worksheet before writing any code. Load the workbook
        and scan all rows/columns to determine the full used range and detect
        multiple table regions. Identify header rows by analyzing the content
        pattern of the first row; do not assume the first row is a header.
        Determine the last data row by scanning from the bottom for the last
        non-empty cell in relevant columns rather than relying on max_row or
        fixed offsets. Normalize text values and numeric keys before filtering
        or grouping. If cells contain formulas, open with data_only=True. Test a
        sample cell to verify row/column references, data types, and function
        compatibility before finalizing.
    - id: value_output_preference
      text: >-
        When writing results to cells, prefer computed numeric or string values
        directly rather than formulas, because the evaluation environment reads
        cell values without recalculating formulas. Use openpyxl to compute
        values in Python and assign them to cells. Reserve formula strings for
        cases where the task explicitly requires a formula as the output and the
        evaluator evaluates formulas. If a formula is necessary, verify that it
        will be computed by testing the file separately.
    - id: column_verification_guidance
      text: >-
        Before hardcoding column indices or row ranges, print the headers of
        the worksheet and the first few data rows to verify column positions and
        content types. For repeated column groups, inspect a sample of data to
        ensure the mapping is correct. Do not rely on assumptions from the
        instruction text alone.

webshop-slots-v2-3:
  input:
    - id: binding_contract
      text: >-
        Treat the target task context as the only binding contract for the new
        skill. Use retrieved examples only as evidence for transferable tactics
        and common failure modes; never treat them as extra requirements and
        never copy their concrete facts, ids, paths, names, or numeric values.
  output:
    - id: reusable_guidance
      text: >-
        Write each section as reusable guidance grounded in the target contract
        and retrieved evidence, so it transfers to the target task. It must not
        restate this specific task, copy a retrieved solution, or give the final
        answer.
    - id: exact_option_matching
      text: >-
        For WebShop option guidance, treat requested variants as hard
        constraints. Tell the agent to select the exact option value when
        available; do not equate nearby colors, sizes, numbered variants,
        mixed/accent variants, or generic labels with a specific requested
        option. If several options partially match, prefer the exact/base option
        and recover rather than buying a partial match.
    - id: preserve_option_state
      text: >-
        When writing procedures, prioritize option-state preservation: select
        required options as soon as they are visible, avoid optional evidence
        detours that risk losing them, and before buying verify the needed
        options are still selected. If the agent inspects Features/Description
        and Buy Now is not valid there, instruct it to return to the product
        page with the valid previous-page action rather than Back to Search.
    - id: calibrated_search_breadth
      text: >-
        Guide search breadth carefully: never introduce a stricter price filter
        than the user stated. Start with discriminative product terms, but if
        full-constraint searches fail, broaden to the core category and inspect
        under-budget candidates because colors, sizes, counts, and
        configurations may appear only as selectable options rather than in
        result titles.
    - id: calibrate_attribute_evidence
      text: >-
        For non-option attributes, calibrate evidence to the live page: if the
        required terms or clear synonyms appear in the title, bullets, Features,
        or Description and the core product, price, and options match, treat the
        attribute as satisfied. Do not demand extra proof or continue searching
        indefinitely unless the page contradicts a hard requirement.
\end{lstlisting}

\begin{lstlisting}[style=yamlconfig,caption={Framework schema and slot budgets from framework.yaml.}]
framework_version: global-skill-framework-v2
description: >-
  Fixed prompt skeleton for skill synthesis. The slot library supplies one
  fixed dataset_identity field plus two optimizable slot categories (input /
  output), each an ordered list of small slots the optimizer may add, delete, or
  modify. Slots render as bullet blocks at the category placeholders below.

fixed_fields:
  - name: dataset_identity
    position: system
    required: true

slot_categories:
  - name: input
    position: user
    render_placeholder: input_policy_block
    max_slots: 8
    max_chars_per_slot: 1200
    responsibility: >-
      How to read the target task and retrieved evidence: the binding target
      contract, treating retrieved examples as evidence not requirements, and
      what must not transfer from retrieved tasks.
  - name: output
    position: user
    render_placeholder: output_policy_block
    max_slots: 8
    max_chars_per_slot: 1200
    responsibility: >-
      How to write the SKILL.md sections as reusable, non-answer guidance:
      when-to-use framing, failure modes, procedures with decision points, and
      the verification checklist.
\end{lstlisting}

\begin{lstlisting}[style=promptbox,caption={Skill synthesis prompt template from framework.yaml.}]
[System Message]
You synthesize reusable benchmark skills into SKILL.md files for one target task.

Core rules:
- The target_task_context is the only binding contract for the new skill.
- Retrieved examples are nearest-neighbor evidence for reusable tactics,
  failure modes, and verification habits. They are not additional target
  requirements.
- Retrieved task context, trajectories, solutions, tests, errors, labels, and
  scores may be used to infer transferable patterns, but not copied as target
  facts.
- Do not directly give the task answer.
- Return only Markdown for SKILL.md.

Dataset identity:
{dataset_identity}

[User Message]
Create a standalone SKILL.md for benchmark {benchmark}.

How to read the target task and retrieved evidence:
{input_policy_block}

How to write the skill:
{output_policy_block}

Write a concise skill that generalizes from the target and retrieved evidence.
It should help a future agent solve this target task, but it must not be a
transcript, a copied retrieved solution, or a direct final answer.

Required Markdown output:
# SKILL.md

## When to use
Describe when this skill applies, the target-specific contract at a high level,
and the evidence the future agent should gather before acting.

## Possible Failure Modes
- List likely mistakes grounded in the target context and retrieved evidence.
- Prefer benchmark-specific, actionable failure modes over generic advice.

## Possible procedures
- Give concise operational guidance that can transfer to the target task.
- Include decision points, recovery checks, and evidence to look for.
- Do not give a direct target solution.

## Verification Checklist
Write a short checklist of concrete checks the future agent should run before
considering the task complete. Cover the target's hard requirements,
output/action format, evaluator-visible constraints, non-copying from
retrieved examples, preservation of unrelated state, and a recovery check for
errors or ambiguous evidence.

Input payload:
{payload_json}
\end{lstlisting}

\begin{lstlisting}[style=promptbox,caption={Per-case critic prompt used during MPO.}]
[System Message]
  You are a per-case critic for a benchmark prompt optimizer. Diagnose one
  current run for one task. Classify whether the outcome is attributable to
  retrieval, the generated skill, execution behavior, or other factors. Return
  only valid JSON.

[User Message]
  Analyze this single optimization case.

  Use the current trajectory, skill, eval signal, task context, skill retrieval
  examples, and ground truth. The skill was synthesized from the
  retrieved trajectories in skill_retrieval: those retrieved_examples are the
  raw material the skill was synthesized from. The transition field reports how
  this task moved from the previous run:
  regression (was correct, now wrong), improvement (was wrong, now correct),
  persistent_failure (wrong in both), persistent_success (correct in both).
  For regression and improvement cases, previous_run holds the prior skill and
  trajectory, and slot_diff lists the slot edits made between the two runs
  (added/removed/modified). For a regression, judge whether a slot_diff edit
  plausibly caused the new failure (for example, the previous skill avoided a
  mistake the new skill introduced); if so, say which edit in the lesson.
  Treat transitions as possibly noisy: an isolated flip can be sampling
  variance rather than a real effect of the edit. current_slot_library lists
  the optimizable slots (id and text, grouped by input/output) that generated
  this run's skill; these are the editable units the optimizer can add, delete,
  or modify, and combined with slot_diff you can reconstruct the previous
  run's library. When your diagnosis traces this case's outcome to one
  specific standing slot -- for example a slot whose guidance plausibly steered
  the skill wrong, or one that plausibly helped -- record that slot's id in
  implicated_slot_id; leave it null when no single existing slot is responsible.
  Do not bias toward adding or toward removing: report only what the evidence
  supports, whichever direction it points. For failed cases, set
  failure_attribution to exactly one of retrieval, skill, execution, or other.
  For succeeded cases, set success_attribution to exactly one of skill,
  execution, or other. Use execution ONLY when the skill's guidance for this
  specific failure was both correct AND concrete/operational, and the agent
  still made an independent low-level error the skill could not reasonably have
  prevented. Do NOT use execution merely because the agent did not follow the
  skill: if the skill only gave generic, vague, or non-operational advice and
  the agent skipped it, attribute to skill, because the actionable lever is
  making that guidance specific and operational. Use retrieval when the
  retrieved examples were missing, misleading, or mismatched in a way that
  plausibly shaped a bad skill. Use skill when the generated skill itself
  plausibly helped or hurt the run, INCLUDING when its guidance was too vague
  or non-operational to change the agent's behavior. Use other for evaluator
  noise, environment issues, impossible/ambiguous tasks, or cases where the
  evidence is insufficient. Prefer concrete ground truth evidence over
  speculative failure explanations when it is available. Attribute claims to
  concrete trajectory, skill, retrieval, and ground truth evidence when supported.
  The lesson field is a single short instruction: for skill-attributed
  failures, say what the prompt should change; for skill-attributed successes,
  say what the prompt should preserve. Keep the lesson benchmark-general and
  reusable across tasks; do not embed this task's specific facts, ids, or
  answers. When the attribution is not skill, set lesson to null or "none". Do
  not propose replacement prompt text and do not copy task answers into
  recommendations.

  Return JSON with exactly these keys:
  {
    "task_id": string,
    "failure_attribution": "retrieval" | "skill" | "execution" | "other" | null,
    "success_attribution": "skill" | "execution" | "other" | null,
    "evidence": list of 1 to 3 concrete grounded observations,
    "lesson": string or null,
    "implicated_slot_id": existing slot id from current_slot_library that this case traces to, or null
  }

  Payload:
  {case_payload_json}
\end{lstlisting}

\begin{lstlisting}[style=promptbox,caption={Slot editor prompt used during MPO.}]
[System Message]
  You are the editor for a benchmark-specific skill-synthesis prompt
  optimizer. You are a restricted slot patcher: use only the current slot
  library, framework constraints, run metrics, and per-case critic attributions.
  Apply a bounded list of atomic slot-level patches and return only valid JSON.

[User Message]
  Create one structured patch as a list of slot operations. The slot library has
  two editable categories, input and output, each an ordered list of small
  single-purpose slots with a stable id and text. Read slot_schema for each
  category's responsibility, max_slots, and max_chars_per_slot. Return between
  1 and 4 operations; each is one of add_slot, delete_slot, or modify_slot.
  Leave every slot you do not touch unchanged. Each operation must target the
  category and slot whose responsibility most directly matches a
  skill-attributed lesson, and must be independently justified by the evidence
  your candidate policy prioritizes.

  Prefer the smallest set of operations that covers the distinct,
  independently-supported themes: one operation per theme, not several refining
  the same idea. Every extra operation makes the next run's transitions harder
  to attribute, so do not pad; if only one theme is well supported, return a
  single operation. Operations apply in order to the same library: a later
  operation sees the effect of earlier ones (for example you may delete_slot
  then add_slot a replacement, or modify two different slots that pull in
  opposite directions so they no longer conflict).

  How to use critic attributions:
  - Treat failure_attribution=skill as the main evidence for what to repair.
  - Treat success_attribution=skill as the main evidence for what to preserve.
  - Treat retrieval/execution/other attributions as cautionary evidence: do not
    edit slots to fix problems the skill prompt likely cannot control.
  - If the most informative lessons are non-skill or say none, make the
    smallest preservation-oriented edit or tighten guardrails against
    overclaiming.
  - last_edit_slot_diff is the slot change made between the previous and
    current run. Each case_critic_results entry carries its transition
    (regression = was correct, now wrong; improvement = was wrong, now correct;
    persistent_failure; persistent_success), so you can read that edit's
    measured effect per task. These transitions come from a stratified sample
    of the run, not the full run, so judge themes, not counts.
  - A critic entry may set implicated_slot_id: the existing slot that case
    traced to. Across the cases, an implicated slot tied to regressions or
    persistent failures is a candidate to modify or delete; one tied to
    improvements or persistent successes is a candidate to preserve or extend.
  - These lessons may point in opposite directions on the same dimension. When
    two conflict, either make the single edit that resolves the conflict or
    scope each rule to its own trigger condition so they no longer collide.
  - Improvements and success-attributed lessons mark guidance worth extending or locking in.

  Operation choice:
  - modify_slot: the responsibility is already covered but the wording is weak, vague, or slightly off.
  - add_slot: a clearly supported rule has no natural home in any existing slot.
  - delete_slot: a slot is actively harmful, or fully redundant with stronger existing guidance.
  - max_slots is counted after all your operations apply in order; if a category
    is already full you must delete_slot before you can add_slot to it within
    the same patch.

  Editing constraints:
  - Keep each edited slot within its category's responsibility; one slot should
    carry one rule, not a catch-all paragraph.
  - Preserve useful existing instructions unless critic lessons identify them as harmful.
  - Do not add specific task answers, retrieved example facts, product names,
    cell values, code constants, room routes, or IDs.
  - Do not move fixed framework rules (dataset_identity, the SKILL.md section
    list, the target-contract framing) into benchmark slots.
  - Prefer concise, operational benchmark-specific text over broad prompt-engineering advice.
  - Keep each added or modified text within the slot_schema max_chars_per_slot
    budget for its category.

  Return JSON with exactly these keys:
  {
    "operations": [
      {
        "operation": "add_slot" | "delete_slot" | "modify_slot",
        "category": "input" | "output",
        "slot_id": existing slot id for delete/modify, or a new stable snake_case id for add,
        "reason": concise explanation grounded in the critic attributions and lessons,
        "text": complete slot text for add/modify, or null for delete
      }
    ]  // 1 to 4 operations, applied in order
  }

  Payload:
  {editor_payload_json}
\end{lstlisting}

\fi

\ifdefined\appendixonly\else
\bibliography{references}
\fi

\end{document}